\documentclass{article}
\usepackage{spconf,amsmath,graphicx,hyperref}

\usepackage{spconf,amsmath,graphicx}
\usepackage{xcolor}
\usepackage{subcaption}
\usepackage{enumitem}
\usepackage{float}
\usepackage[utf8]{inputenc}
\usepackage[T1]{fontenc}
\usepackage{url}
\usepackage{booktabs}
\usepackage{amssymb}
\usepackage{microtype}
\usepackage{multirow}
\usepackage{cleveref}

\graphicspath{{ptbxl_analysis/}}

\extrafloats{100}
\def\z{{\mathbf z}}
\newcommand{\bvit}{\textsc{bViT}}
\newcommand{\vit}{\textsc{ViT}}
\newcommand{\Lfive}{\textsc{ptb-5}}
\newcommand{\Lfortyfour}{\textsc{ptb-44}}

\title{Same Path, Different Walk: A Mechanistic Comparison of Looped\\
and Stacked Transformer Encoders on 12-Lead ECG}

\name{Pawel Olszowiec$^{1*}$
\hspace{0.1em} Michal Byra$^{1,2}$ 
\hspace{0.1em} Grzegorz Gruszczynski$^{1}$  
\hspace{0.1em} Grzegorz Stefanski$^{1}$  
\hspace{0.1em} Alberto Presta$^{1}$ }
\address{$^{1}$Samsung AI Center, Warsaw, Poland  \quad \ $^{2}$IFTR, Polish Academy of Sciences, Warsaw, Poland}

\begin{document}
\maketitle
\begin{abstract}
Recurrent Transformers reusing their weights rather than stacking $L$ distinct layers are becoming widely adopted due to their parameter efficiency \cite{byra2026bvitinvestigatingsingleblockrecurrence, zhang2022minivit, shen2022sliced}. 
However, the exact representational and dynamical differences between looped and stacked architectures remain uncharacterized. 
This paper presents a controlled study on the example of \bvit{} model \cite{byra2026bvitinvestigatingsingleblockrecurrence} applying one weight-tied block $L$ times. We train two models: \bvit{} and standard \vit{} \cite{dosovitskiy2020image} on 12-lead electrocardiogram (ECG) classification tasks from the PTB-XL dataset under identical training protocols. 
Despite an $8.9\times$ parameter reduction, \bvit{} achieves accuracy parity with \vit{}. 
Geometric similarity metrics demonstrate that both architectures construct comparable latent representations in an equivalent canonical order. 
Crucially, their dynamics differ: \bvit{} exhibits smaller step sizes and inter-patient sensitivity, as well as near-neutral behavior away from the data manifold, whereas \vit{} exhibits collapsing dimensionality of representations and out-of-distribution feature expansion.
\end{abstract}
\begin{keywords}
Robust and trustworthy machine learning, Explainable and interpretable machine learning, Transformers, ECG Signal Processing, Automated Cardiac Diagnostics
\end{keywords}

\section{Introduction} \label{sec:intro} Standard Vision Transformer (\vit{}) architectures stack $L$ distinct parameterized layers to process sequential representations. In contrast, looped Transformers iteratively apply a \emph{single} weight-tied block $L$ times~\cite{dehghani2018universal,lan2019albert}, matching deep baselines across vision benchmarks with drastic parameter reductions~\cite{zhang2022minivit, byra2026bvitinvestigatingsingleblockrecurrence}. Prior studies have firmly established empirical accuracy parity for looped architectures when given sufficient capacity~\cite{gruszczynski2026training}.
However, in safety-critical clinical applications—such as continuous 12-lead electrocardiogram (ECG) (see e.g. \cite{ribeiro2020automatic, hannun2019cardiologist, mehari2022self}) monitoring on resource-constrained bedside hardware~\cite{wagner2020ptb, strodthoff2020deep}—classification accuracy alone does not capture operational safety. A fundamental open question remains: \emph{Do looped and stacked Transformers arrive at identical internal representations through equivalent dynamical mechanisms, or does weight tying impose distinct inductive biases?}
As shown in \Cref{tab:headline}, \bvit{} matches \vit{} on 5-class ECG diagnosis (\Lfive{}) and on 44-subclass diagnosis (\Lfortyfour{}) while utilizing $8.9\times$ fewer parameters. Through geometric and dynamical systems analyses, this work investigates the internal mechanics underlying this parity. 
Alternative weight-sharing designs such as MiniViT~\cite{zhang2022minivit} and Sliced Recursive Transformers \cite{shen2022sliced} introduce per-step operator variation -through per-layer parameters or block slicing - that destroys the time-invariance on which the dynamical analysis rests. 
\bvit{} is the minimal construction allowing for controlled comparison that lets the paper attribute all differences to weight sharing alone.
\begin{figure}[t]
  \centering
  \begin{subfigure}{0.49\columnwidth}
    \includegraphics[width=\textwidth]{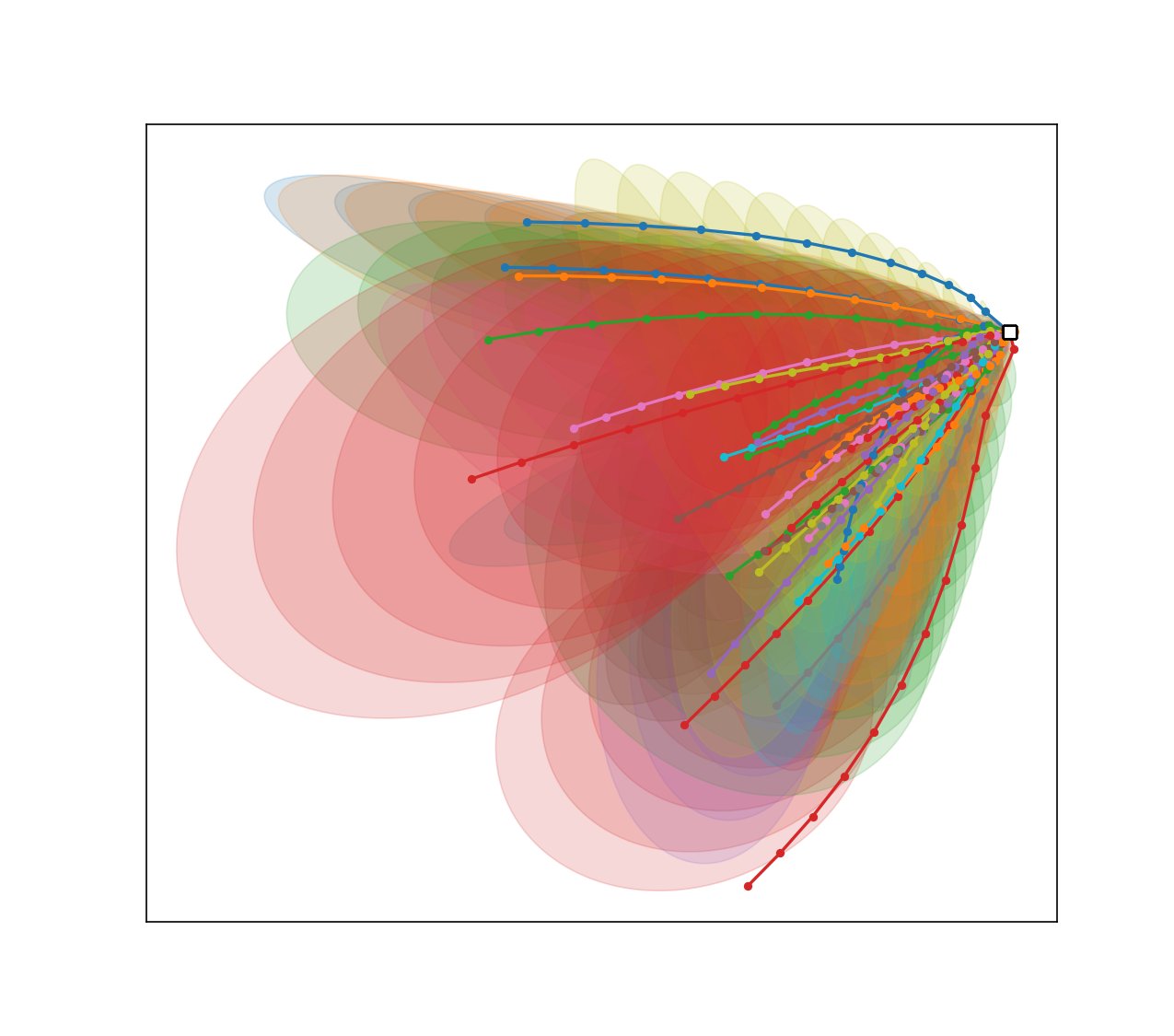}
    \caption{\bvit{}.}
  \end{subfigure}\hfill
  \begin{subfigure}{0.49\columnwidth}
    \includegraphics[width=\textwidth]{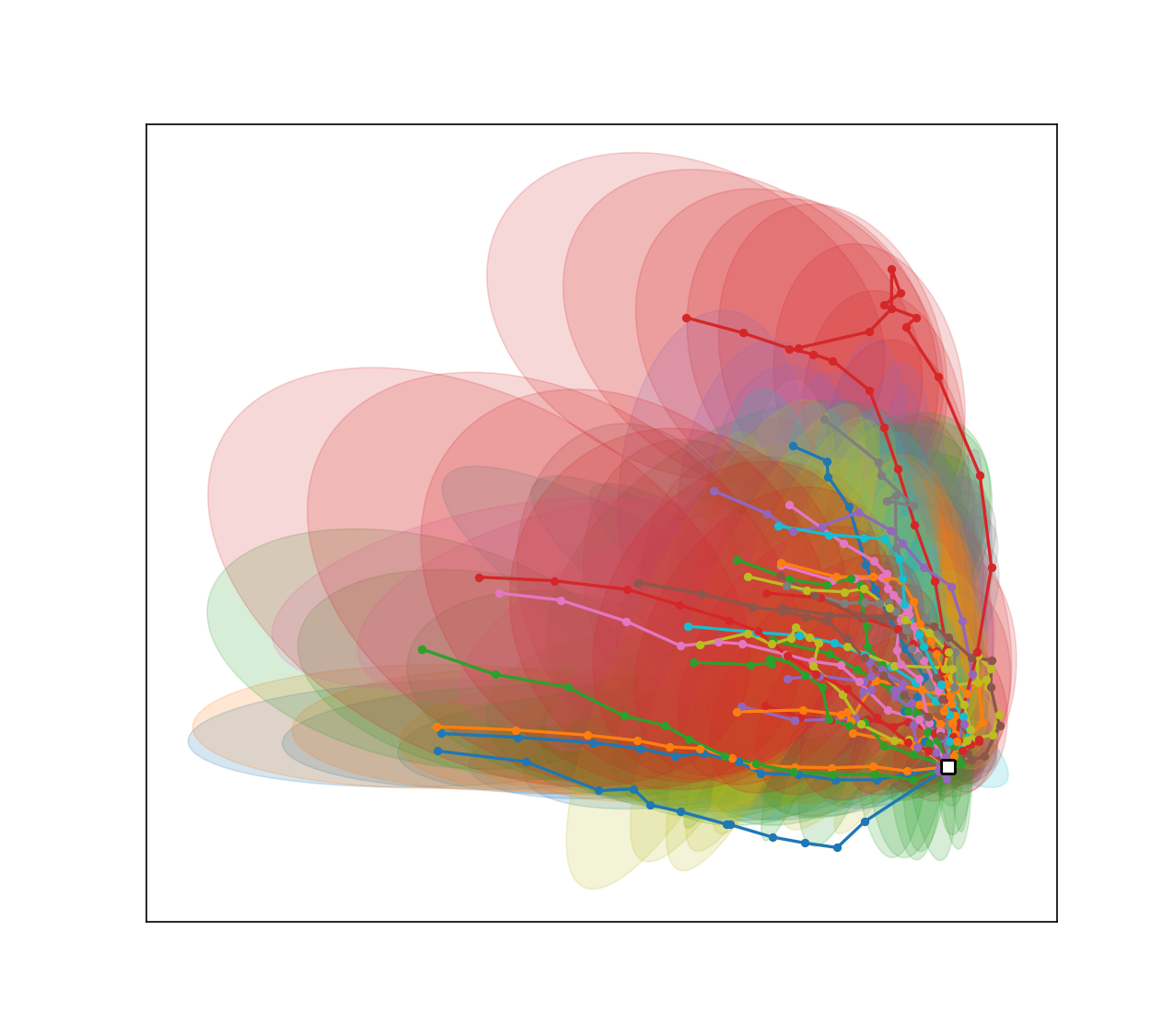}
    \caption{\vit{}.}
  \end{subfigure}
  \caption{Joint PCA of class trajectories with $\sigma=1$ covariance ellipses. 
  \bvit{} follows smooth radial trajectories, \vit{} shows piecewise directional shifts.}
  \label{fig:fans}
\end{figure}
\subsection{Scope \& Contributions}
This study is a controlled mechanistic investigation, not a competitive benchmarking effort. We fix the data distribution, patch tokenization, and optimization pipeline to isolate the architectural effect of weight sharing. Our core findings are:
\begin{enumerate}[leftmargin=1.4em,itemsep=1pt,label=\textbf{C\arabic*.}]
  \item \textbf{Monotonic Dynamical Stability:} \bvit{} exhibits smooth state transformations and monotonically decreasing updates, whereas \vit{} maintains elevated step variance and dimensionality collapse (\Cref{sec:c1}).
  \item \textbf{Latent Representation Equivalence:} \bvit{} and \vit{} construct geometrically aligned latent representations across depth in the same canonical order (\Cref{sec:c2}).
  \item \textbf{Deployment Robustness:} \bvit{} demonstrates 
  lower inter-patient Jacobian variance and preserves geometric cluster separability on out-of-distribution inputs (\Cref{sec:c3}).
\end{enumerate}

\begin{table}[t]
\centering
\caption{Parameter counts and test performance. \Lfive{} scored by accuracy (\%); \Lfortyfour{} by macro-AUC.}
\label{tab:headline}
\resizebox{\columnwidth}{!}{
\begin{tabular}{lcccc}
\toprule
 & \multicolumn{2}{c}{\Lfive{} (Acc.\ \%)} & \multicolumn{2}{c}{\Lfortyfour{} (AUC)} \\
\cmidrule(lr){2-3}\cmidrule(lr){4-5}
 & \bvit{} & \vit{} & \bvit{} & \vit{} \\
\midrule
Encoder body  & $\mathbf{789\text{K}}$ & $9.47\text{M}$ & $\mathbf{789\text{K}}$ & $9.47\text{M}$ \\
Total params  & $\mathbf{1.10\text{M}}$ & $9.78\text{M}$ & $\mathbf{1.11\text{M}}$ & $9.79\text{M}$ \\
Reduction     & \multicolumn{2}{c}{$\mathbf{8.9\times}$} & \multicolumn{2}{c}{$\mathbf{8.8\times}$} \\
\midrule
\textbf{Test Performance} & $\mathbf{67.11}\pm0.32$ & $67.06\pm0.47$ & $\mathbf{0.8472}\pm0.0051$ & $0.8381\pm0.0062$ \\
\bottomrule
\end{tabular}
}
\end{table}

\section{Experimental Setup}
\label{sec:setup}
\subsection{Architectures: bViT vs ViT} Both models~\cite{byra2026bvitinvestigatingsingleblockrecurrence, dosovitskiy2020image} process $12\times1000$ ECG windows as ten non-overlapping 100-sample patches, projected to $d=256$ dimensional latent space, with a \textsc{cls} token, through 12 steps. They differ only in parameter sharing: \vit{} uses 12 distinct blocks ($9.47\text{M}$), \bvit{} reuses one ($789\text{K}$).
A block is the standard pre-norm Transformer encoder ~\cite{vaswani2017attention}, with $H=8$ attention heads, MLP expansion factor 4, dropout $0.2$
\begin{equation}
\label{eq:block}
\begin{aligned}
  \mathbf{u} &= \mathbf{h} + \mathrm{MHSA}\!\left(\mathrm{LN}(\mathbf{h})\right), \\
  B(\mathbf{h}) &= \mathbf{u} + \mathrm{MLP}\!\left(\mathrm{LN}(\mathbf{u})\right).
\end{aligned}
\end{equation}
Let $z_t$ denote hidden states $t$ block applications, $z_0$ the embedded
input, and $\z_t\in\mathbb{R}^{256}$ the \textsc{cls} row of $z_t$. Then for
$t=1,\dots,12$,
\begin{equation}
\label{eq:models}
  \textbf{\bvit{}:}\;\; z_t = B(z_{t-1}),
  \qquad
  \textbf{\vit{}:}\;\; z_t = B_t(z_{t-1}).
\end{equation}
The initial projection ($\z_0 \to \z_1$) represents static initialization and is excluded from iterative dynamical statistics.

\subsection{Data} We evaluate on 12-lead ECG (PTB-XL~\cite{wagner2020ptb}) because a patch Transformer over a multichannel and quasi-periodic waveform is naturally cast as a learned filter bank or iterative cascade, for which gain, stability, and contraction---the standard concerns of signal processing---are the right diagnostic quantities. 
The dataset is small enough so that both \bvit{} and \vit{} converge to their metric plateau under a single recipe, so dynamical differences are not confounded with under-training. We use the recommended split~\cite{strodthoff2020deep} (folds 1--8 train, 9 val, 10 test; $17,418$/$2,183$/$2,198$) with per-channel normalisation. \textbf{\Lfive{}} (5 superclasses, single-label, cross-entropy, accuracy) and \textbf{\Lfortyfour{}} (44 subclasses, multi-label, BCE, macro-AUC) differ in loss, metric and difficulty - they serve as internal replication. 
All the figures and statistics present data, unless stated otherwise, obtained on \Lfortyfour{} test set from 20 different training runs of \bvit{} and \vit{}. 

\subsection{Training the two models} We train \bvit{} and \vit{} from scratch for $100$ epochs with AdamW~\cite{loshchilov2017decoupled}, batch
size $128$, initial learning rate $5\times10^{-4}$ cosine-decayed to $10^{-5}$,
10 warm-up epochs, weight decay $0.1$ and gradient clipping $1.0$, no weight averaging. 
Augmentation is applied to the training split only, each
transform drawn independently with probability $0.5$: Gaussian noise
($\sigma\leq0.03$), random circular time shift ($\pm10\%$), amplitude scaling
($0.8$--$1.2\times$), single-lead channel dropout and random time masking.

\section{DYNAMICAL SMOOTHNESS AND CAPACITY (C1)} \label{sec:c1} We characterize state evolution $\z_1,\dots,\z_{12}$ via four dynamical properties to answer the following questions: does the state settle, how much of the available space does it occupy, and is a perturbation amplified or damped?

\textbf{Class Geometric Trajectories:} (\Cref{fig:fans}) Per-class averages traced across depth reveal whether classes separate along stable directions or are repeatedly rearranged.

\begin{figure}[t]
  \centering
  \begin{subfigure}{0.99\columnwidth}
    \includegraphics[width=\textwidth]{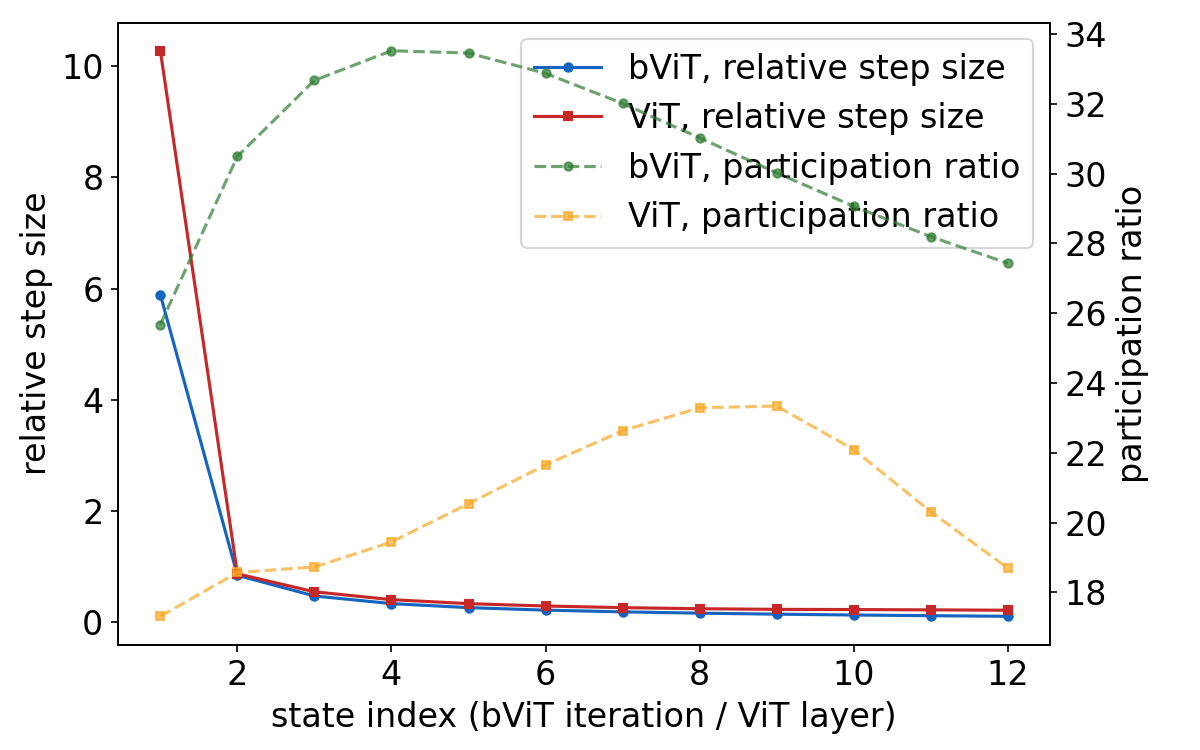}
  \end{subfigure}
  \caption{Trajectory statistics. \bvit{} performs smaller relative latent steps than \vit{}   
  and holds a wider representation at every depth. }
  \label{fig:dyn}
\end{figure}

\textbf{Update Magnitude:} (\Cref{fig:dyn}a) The relative step size $RSS= \lVert\z_t-\z_{t-1}\rVert/\lVert\z_{t-1}\rVert$ measures the fraction of the current representation each block rewrites. A sequence decreasing to 0 indicates settling in latent space, while others indicate that final blocks are still revising a latent representation about to be classified.

\textbf{Effective dimensionality:} (\Cref{fig:dyn}b) Let $\lambda_1,\dots,\lambda_{256}$ be the eigenvalues of the covariance matrix of $\z_t$ over the test subset. The participation ratio \cite{johnston2023abstract} $\mathrm{PR}(\z_t) = \frac{\left(\sum_i \lambda_i\right)^2}{\sum_i \lambda_i^2}$ counts how many directions carry comparable variance — i.e. the effective dimensionality of the representation at depth $t$. A collapsed representation has less room to separate classes and distorts similarity scores.

\begin{figure}[ht]
  \centering
  \includegraphics[width=0.99\columnwidth]{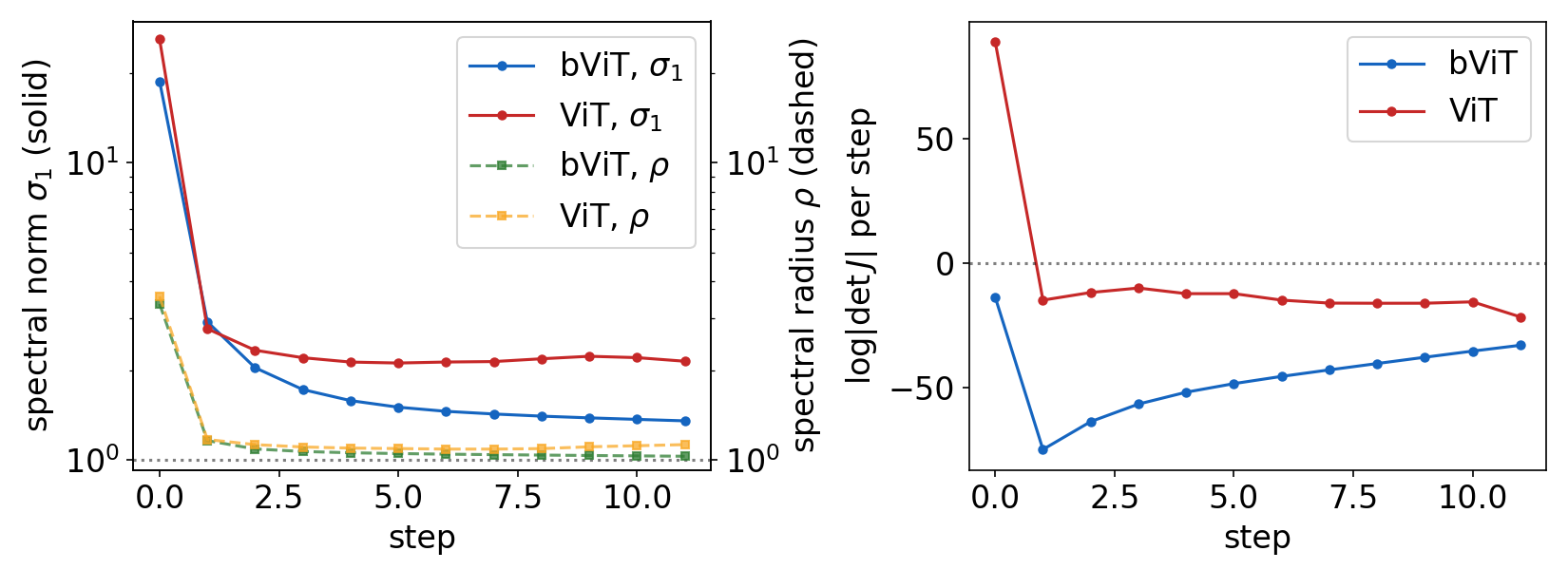}
  \caption{Single-step Jacobian. 
  \textbf{Left:} spectral norm (solid) and spectral radius (dashed). \bvit{} has smaller spectral norm and spectral radius than \vit{} and so it is more stable model.  
  \textbf{Right:} $\log|\det J_t|$. Every \bvit{} step contracts; \vit{} expands on its first step.}
  \label{fig:jac}
\end{figure}

\textbf{Linearised step response:} (\Cref{fig:jac}) 
For one \bvit{} iteration or one \vit{} layer we take
$J_t=\partial\z_t/\partial\z_{t-1}$ and extract three quantities. 
The spectral norm $\lVert J_t\rVert_2$ (largest singular value - $\sigma_1$) is the worst-case amplification of a perturbation in a single step. 
The spectral radius $\rho(J_t)$ (largest eigenvalue magnitude) governs repeated application: above $1$
perturbations grow under iteration, below $1$ they decay. 
And $\log|\det J_t|$ is the volume change, negative when a cloud of nearby recordings is contracted by
the step and positive when it is expanded. 
$J_t$ for both models are estimated by power iteration and stochastic Lanczos quadrature~\cite{hutchinson1989stochastic,ubaru2017fast}. \Cref{tab:summary} collects results at both label levels, \Lfive{} and \Lfortyfour{}.

\begin{table}[t]
\centering
\caption{Summary at both label levels. \emph{Sensitivity, mean} and \emph{spread}: mean and std of $\rho(J)$ over test recordings. \emph{Volume change}: $\sum_t\log|\det J_t|$ (negative $=$ contraction). Bold marks the better value; sensitivity mean is unmarked (because $\rho\approx1$ is neutral, not good).}
\label{tab:summary}
\resizebox{\columnwidth}{!}{
\begin{tabular}{lcccc}
\toprule
 & \multicolumn{2}{c}{\Lfive{}} & \multicolumn{2}{c}{\Lfortyfour{}} \\
\cmidrule(lr){2-3}\cmidrule(lr){4-5}
 & \bvit{} & \vit{} & \bvit{} & \vit{} \\
\midrule
Effective dimensionality, peak  & $\mathbf{14.8}$ & $7.1$ & $\mathbf{33.8}$ & $23.4$ \\
Effective dimensionality, end   & $\mathbf{6.4}$ & $3.8$ & $\mathbf{27.4}$ & $18.6$ \\
Relative step size, last        & $\mathbf{0.118}$ & $0.209$ & $\mathbf{0.104}$ & $0.212$ \\
\midrule
Sensitivity, mean      & $1.042$ & $1.528$ & $1.044$ & $1.430$ \\
Sensitivity, spread    & $\mathbf{0.014}$ & $0.478$ & $\mathbf{0.008}$ & $0.160$ \\
Volume change          & $\mathbf{-164}$ & $+56$ & $\mathbf{-545}$ & $-74$ \\
\bottomrule
\end{tabular}}
\end{table}

\section{REPRESENTATIONAL CONVERGENCE (C2)} \label{sec:c2}
\begin{figure}[t]
  \centering
  \begin{subfigure}{0.36\columnwidth}
    \includegraphics[width=\textwidth]{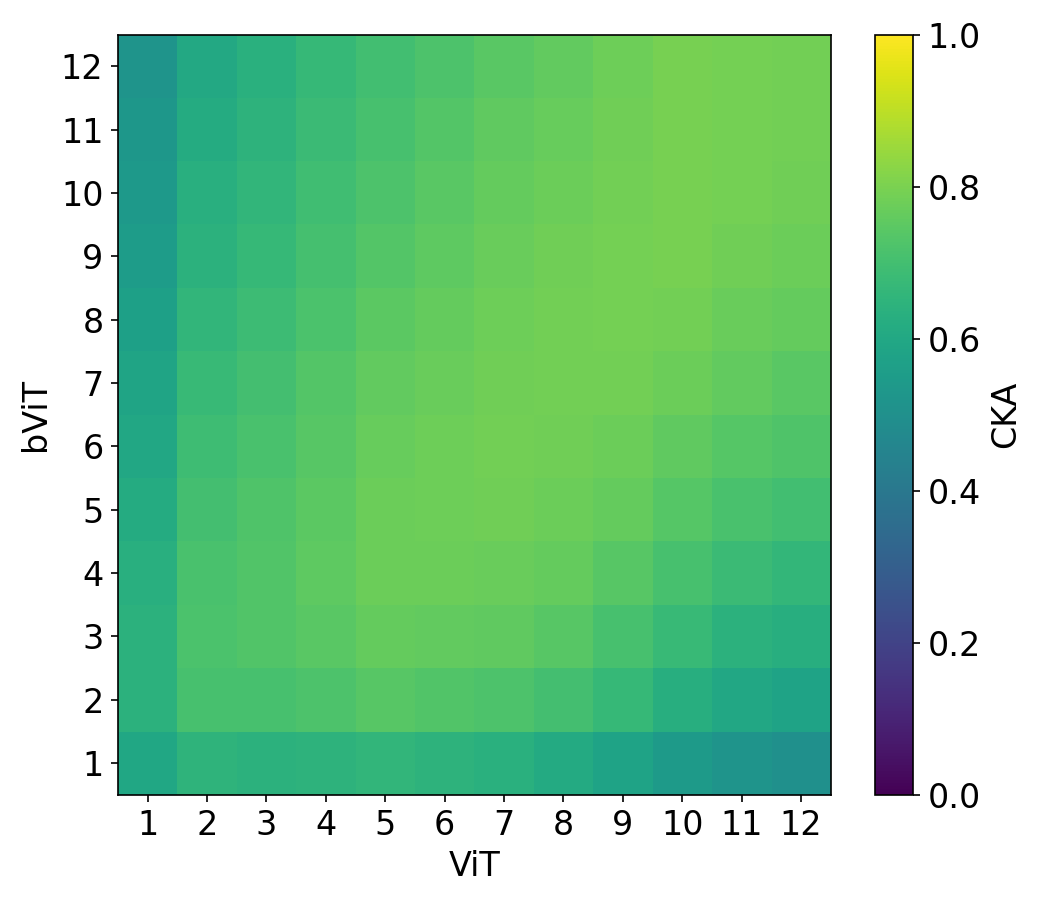}
    \caption{Linear CKA}
  \end{subfigure}\hfill
  \begin{subfigure}{0.63\columnwidth}
    \includegraphics[width=\columnwidth]{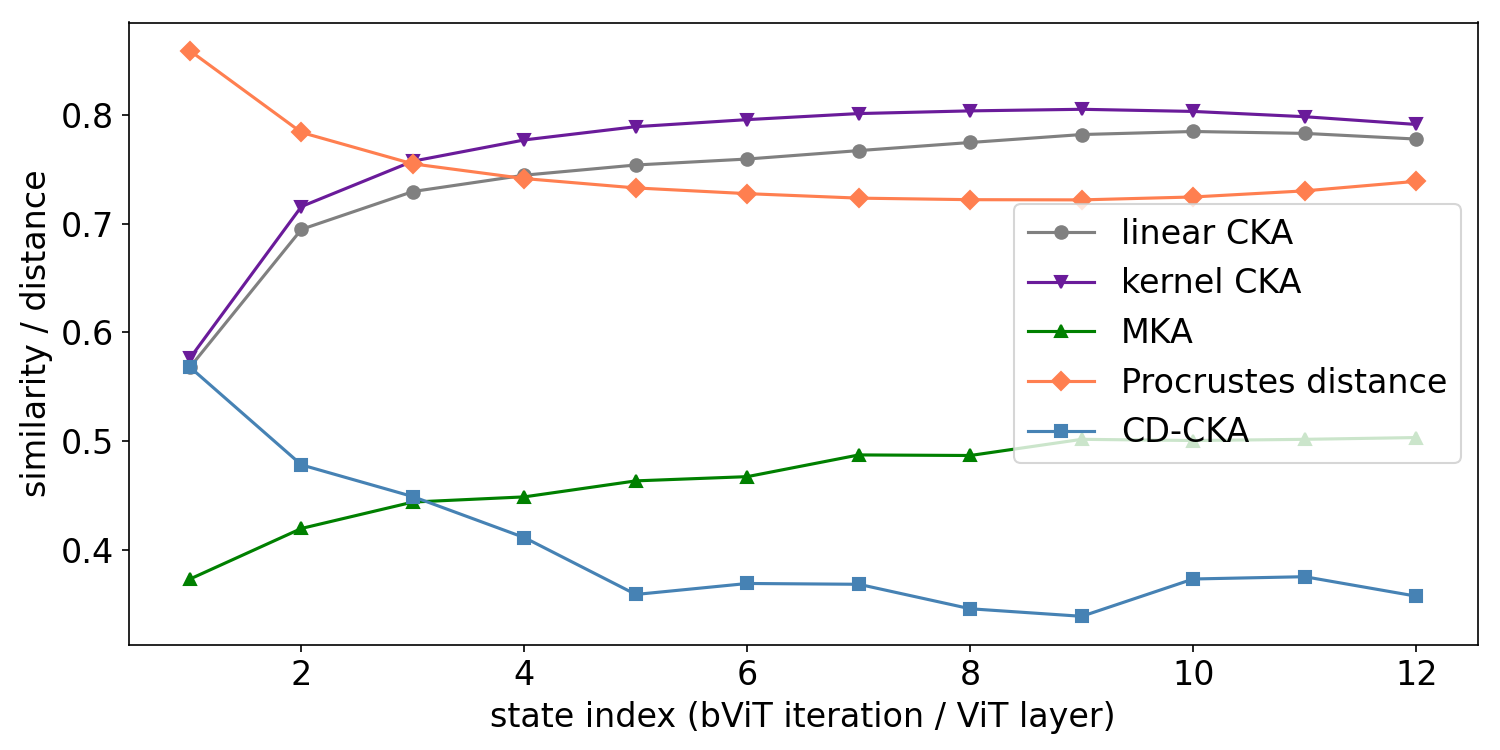}
    \caption{Five similarity measures}
  \end{subfigure}
  \caption{The two models take \textbf{different routes} to comparable destinations. 
  (a) Linear CKA, \bvit{} iterations (rows) vs.\ \vit{} layers (columns). 
  (b) Five measures along the matched diagonal. Three measures initially rise with depth and reach plateau. The CD-CKA does not - a consequence of weight sharing.}  
  \label{fig:cka}
\end{figure}

We ask whether, at equal accuracy, iteration $t$ of \bvit{} arrives at a representation geometrically comparable to layer $t$ of \vit{}. If so, weight sharing changes the trajectory without changing the similarity of destination. If not, the two models solve the task by different means. 
Our primary instrument is linear CKA~\cite{pmlr-v97-kornblith19a}, which compares the similarity structure two representations induce over the same inputs and is invariant to rotation and isotropic rescaling. Because linear CKA is dominated by the highest-variance directions \cite{davari2022reliability}, and \Cref{sec:c1} showed both encoders concentrate onto few directions, we supplement it with four additional similarity measures: CKA based on Gaussian RBF kernel (captures nonlinear structure) \cite{pmlr-v97-kornblith19a}, MKA — Manifold-Approximated Kernel Alignment (based on sparse, directed Nearest Neighbor graphs) \cite{islam2025kernel}, a Procrustes distance (lower = more similarity) \cite{schonemann1966generalized}, and a linear CD-CKA (Contrastive-Difference) which compares increments $\z_t-\z_{t-1}$ and asks if \bvit{} and \vit{} move alike rather than just arrive alike. 
Only the trend with depth should be read across measures, since each normalizes differently (\Cref{fig:cka}).
Cross-model CKA (\Cref{fig:cka}a) rises from $0.55$ to $0.78$ across depth, with layer-to-iteration mapping in the same order, ($[3,5,5,6,7,8,9,9,9,10,10,10]$): later iterations match later layers, so the loop never revisits earlier stages. 
The highest CKA similarity occurs between \vit{} layers $3$--$10$, indicating that the majority of the latent transformation happens there, while the remaining layers make minor contributions.
The pattern is identical on \Lfive{}.
Three of five similarity measures rise with depth. On the other hand the CD-CKA measure does not, revealing the signature of weight sharing, caused by \bvit{}'s self-similarity rising from $0.65$ to $0.99$ since the same block is reapplied to an increasingly settled state, while \vit{}'s stays between $0.22$ and $0.43$. 
The weight sharing constrains the trajectory so differences appear in dynamics (\Cref{sec:c1}) and robustness (\Cref{sec:c3}), not in the final representations. 

\section{PERTURBATION AND OOD ROBUSTNESS (C3)} \label{sec:c3} Clinical translation requires models to remain predictable across diverse patient cohorts and unseen recording artifacts.
We evaluate this along two axes: Inter-Patient Sensitivity Homogeneity and Far-Field OOD Dynamics.
\Cref{sec:c1,sec:c2} established that the \bvit{} and \vit{} reach comparable representations by different routes, at equal accuracy. Neither result speaks to behaviour when the input is not drawn from the test fold — the operative question for a deployed ECG model. We approach it from two directions: how far the models's behaviour varies between patients, and what it does to inputs far outside the training distribution.
\subsection{Inter-Patient Sensitivity Homogeneity} \label{sec:inter-patient} 
\begin{figure}[t]
   \centering
  \includegraphics[width=0.9\columnwidth]{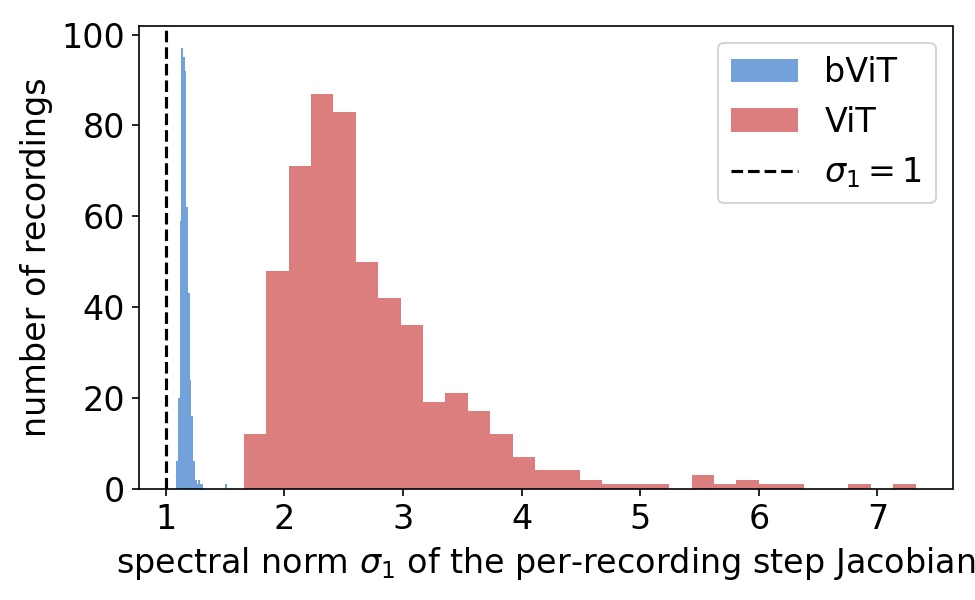}
  \caption{Histogram of local sensitivity. \bvit{}: narrow spike at $1.044\pm0.008$. \vit{}: broad, $2.430\pm0.160$, tail exceeding $1.8$. Spread is $21\times$ smaller for \bvit{} ($34\times$ on \Lfive{}).}
  \label{fig:jachist}
\end{figure}

For each test recording we form the exact Jacobian $J$ of a single step and take its spectral norm $\sigma_1(J)$, the local gain: the factor by which a small change in that patient's representation can be amplified by one forward pass. The location of the resulting distribution (\Cref{fig:jachist}) indicates whether a single step has any potential expansive direction ($\sigma_1>1$), is neutral ($\sigma_1\approx1$), or strictly contractive ($\sigma_1<1$). None of the models is strictly contractive. The narrow distribution of \bvit{} means that a robustness measurement on the test cohort is informative about other cohorts and is model-specific; \vit{} does not have this property.
\subsection{Far-Field OOD Dynamics} 
\begin{figure}[t]
  \centering
  \includegraphics[width=0.99\columnwidth]{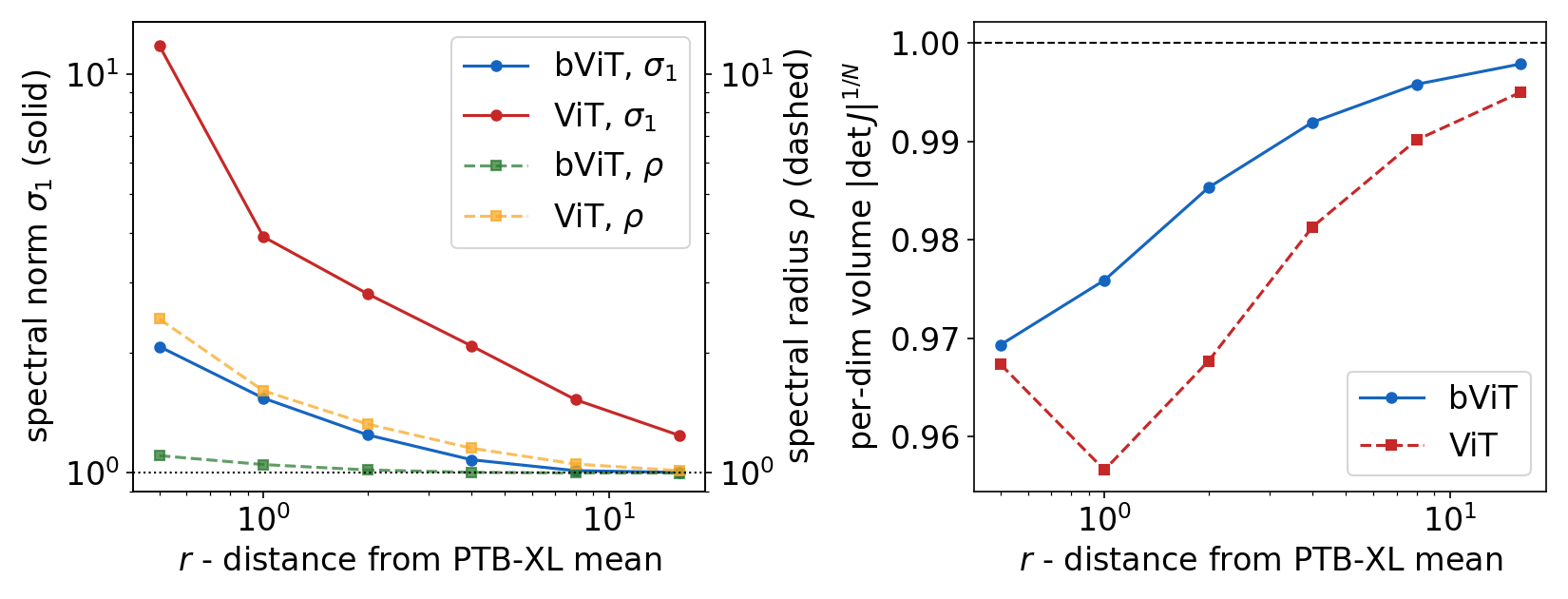}
  \caption{Single-step behaviour vs.\ distance from training data, \Lfortyfour{}. $r\approx1$ is on-data (stars), $r=8$ is far OOD. \textbf{Left:} gain (spectral norm, left; spectral radius, right). \textbf{Right:} per-dimension volume factor $|\det J|^{1/N}$, below one for both. Both are most active on-data; the difference is in the far field.}
  \label{fig:ffsweep}
\end{figure}

\begin{figure}[t]
  \centering
  \includegraphics[width=0.97\columnwidth]{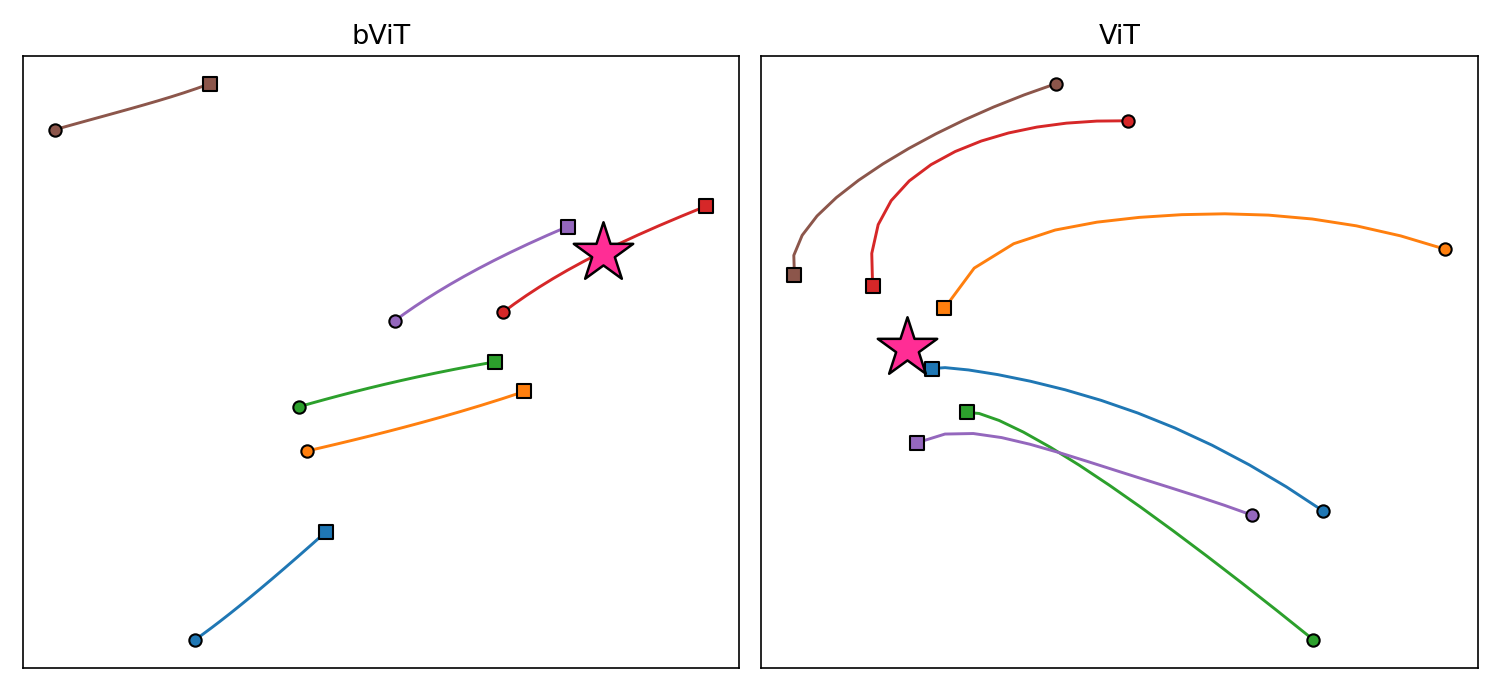}
  \caption{Do distinct OOD groups remain distinct? Six probe clusters at $r=8$. Joint PCA projection of cluster-center paths over twelve steps, squares $=$ start, circles $=$ step 12.  The pink star marks the training-data mean $\mu$ projected into that frame. 
  \bvit{} is near-isometric OOD; \vit{} is non-linearly expansive. 
  }
  \label{fig:ffclusters}
\end{figure}

An unfamiliar patient population can be viewed in latent space as a set of points lying in different direction or being more distant from the training data than the test set. We probe each model at controlled distances from the training mean $\mu$, expressed in units of the test data spread: $s = \underset{z \in S}{\mathbb{E}} \lVert\z-\mu\rVert$, where $S$ is the set of embeddings corresponding to test set samples. Probe points at $r\cdot s$ from $\mu$ along random directions place $r\approx1$ among real recordings and $r=8$ far outside the training distribution. We track three quantities as $r$ grows: single-step gain (\Cref{fig:ffsweep}), boundedness under iteration (\Cref{fig:ffclusters}), and group separability---the inter-cluster distance divided by intra-cluster distance. \bvit{} loses $0.8\%$ ($28.6\to28.3$ across iterations), while \vit{} loses $5.6\%$ ($28.6\to27.0$ across layers), i.e.\ seven times more than \bvit{}.
The desirable outcome from single-step gain is the neutral one — gain near $1$, points that remain in place, groups that stay separated — indicating the model's cautious behaviour - it neither amplifies nor destroys structure it was never trained on. We note that \bvit{} has this property, while \vit{} does not.
\subsection{Implications for deployment} An unfamiliar cohort is a set of out-of-distribution points in latent space. If the model preserves distances in that regime, the cohort's structure survives the forward pass and the feature scale is input-independent. This supports cluster separation and hence performance on out-of-distribution patients, and makes the representation a better initialisation for transfer and fine-tuning.

\section{Conclusion}\label{sec:concl} Replacing \vit{}'s twelve blocks by \bvit{}'s one block applied twelve times incurs no accuracy loss while reducing parameters $8.9\times$. Weight sharing preserves the destination but alters the trajectory, and the trajectory governs deployment behaviour. \bvit{} takes monotonically decreasing steps, contracts volume at every step, and varies $21$--$34\times$ less in inter-patient sensitivity.
The practical reading is that once two models match on accuracy, accuracy has ceased to be informative, and the weight-sharing model is the safer default for deployment - it treats every patient alike and remains neutral on data it was never shown. 
\section{Limitations}\label{sec:limits} 
For controlled comparison, and isolating a single architectural variable---weight sharing in Transformers, we hold the training recipe fixed, for multiple runs from different initializations. We use one corpus, while the two label levels give us internal replication across distinct learning problems.

{
\small
\subsection*{Funding acknowledgements}
This work was supported by Samsung AI Center, Warsaw.

\subsection*{Compliance with ethical standards}
The authors have no relevant conflicts of interest to disclose.

\bibliographystyle{IEEEbib}
\bibliography{strings,refs}

@article{wagner2020ptb,
  title={PTB-XL, a large publicly available electrocardiography dataset},
  author={Wagner, Patrick and Strodthoff, Nils and Bousseljot, Ralf-Dieter and Kreiseler, Dieter and Lunze, Fatima I and Samek, Wojciech and Schaeffter, Tobias},
  journal={Scientific data},
  volume={7},
  number={1},
  pages={154},
  year={2020},
  publisher={Nature Publishing Group UK London}
}

@inproceedings{islam2025kernel,
  title={Kernel Alignment using Manifold Approximation},
  author={Islam, Mohammad Tariqul and Liu, Du and Sarkar, Deblina},
  booktitle={Second Workshop on Representational Alignment at ICLR 2025}
}

@article{dosovitskiy2020image,
  title={An image is worth 16x16 words: Transformers for image recognition at scale},
  author={Dosovitskiy, Alexey and Beyer, Lucas and Kolesnikov, Alexander and Weissenborn, Dirk and Zhai, Xiaohua and Unterthiner, Thomas and Dehghani, Mostafa and Minderer, Matthias and Heigold, Georg and Gelly, Sylvain and others},
  journal={arXiv preprint arXiv:2010.11929},
  year={2020}
}

@article{dehghani2018universal,
  title={Universal transformers},
  author={Dehghani, Mostafa and Gouws, Stephan and Vinyals, Oriol and Uszkoreit, Jakob and Kaiser, {\L}ukasz},
  journal={arXiv preprint arXiv:1807.03819},
  year={2018}
}

@article{lan2019albert,
  title={Albert: A lite bert for self-supervised learning of language representations},
  author={Lan, Zhenzhong and Chen, Mingda and Goodman, Sebastian and Gimpel, Kevin and Sharma, Piyush and Soricut, Radu},
  journal={arXiv preprint arXiv:1909.11942},
  year={2019}
}

@article{strodthoff2020deep,
  title={Deep learning for ECG analysis: Benchmarks and insights from PTB-XL},
  author={Strodthoff, Nils and Wagner, Patrick and Schaeffter, Tobias and Samek, Wojciech},
  journal={IEEE journal of biomedical and health informatics},
  volume={25},
  number={5},
  pages={1519--1528},
  year={2020},
  publisher={IEEE}
}

@article{hannun2019cardiologist,
  title={Cardiologist-level arrhythmia detection and classification in ambulatory electrocardiograms using a deep neural network},
  author={Hannun, Awni Y and Rajpurkar, Pranav and Haghpanahi, Masoumeh and Tison, Geoffrey H and Bourn, Codie and Turakhia, Mintu P and Ng, Andrew Y},
  journal={Nature medicine},
  volume={25},
  number={1},
  pages={65--69},
  year={2019},
  publisher={Nature Publishing Group US New York}
}

@article{ribeiro2020automatic,
  title={Automatic diagnosis of the 12-lead ECG using a deep neural network},
  author={Ribeiro, Ant{\^o}nio H and Ribeiro, Manoel Horta and Paix{\~a}o, Gabriela MM and Oliveira, Derick M and Gomes, Paulo R and Canazart, J{\'e}ssica A and Ferreira, Milton PS and Andersson, Carl R and Macfarlane, Peter W and Meira Jr, Wagner and others},
  journal={Nature communications},
  volume={11},
  number={1},
  pages={1760},
  year={2020},
  publisher={Nature Publishing Group UK London}
}

@article{mehari2022self,
  title={Self-supervised representation learning from 12-lead ECG data},
  author={Mehari, Temesgen and Strodthoff, Nils},
  journal={Computers in biology and medicine},
  volume={141},
  pages={105114},
  year={2022},
  publisher={Elsevier}
}

@article{davari2022reliability,
  title={Reliability of cka as a similarity measure in deep learning},
  author={Davari, MohammadReza and Horoi, Stefan and Natik, Amine and Lajoie, Guillaume and Wolf, Guy and Belilovsky, Eugene},
  journal={arXiv preprint arXiv:2210.16156},
  year={2022}
}

@InProceedings{pmlr-v97-kornblith19a,
  title = 	 {Similarity of Neural Network Representations Revisited},
  author =       {Kornblith, Simon and Norouzi, Mohammad and Lee, Honglak and Hinton, Geoffrey},
  booktitle = 	 {Proceedings of the 36th International Conference on Machine Learning},
  pages = 	 {3519--3529},
  year = 	 {2019},
  editor = 	 {Chaudhuri, Kamalika and Salakhutdinov, Ruslan},
  volume = 	 {97},
  series = 	 {Proceedings of Machine Learning Research},
  month = 	 {09--15 Jun},
  publisher =    {PMLR},
  url = 	 {https://proceedings.mlr.press/v97/kornblith19a.html}
}

@article{ubaru2017fast,
  title={Fast estimation of tr(f(A)) via stochastic Lanczos quadrature},
  author={Ubaru, Shashanka and Chen, Jie and Saad, Yousef},
  journal={SIAM Journal on Matrix Analysis and Applications},
  volume={38},
  number={4},
  pages={1075--1099},
  year={2017},
  publisher={SIAM}
}

@article{hutchinson1989stochastic,
  title={A stochastic estimator of the trace of the influence matrix for Laplacian smoothing splines},
  author={Hutchinson, Michael F},
  journal={Communications in Statistics-Simulation and Computation},
  volume={18},
  number={3},
  pages={1059--1076},
  year={1989},
  publisher={Taylor \& Francis}
}

@article{gruszczynski2026training,
  title={Training Crossroads for Recurrent Vision Transformers: Recurrence, Neural ODEs, and Deep Supervision},
  author={Gruszczynski, Grzegorz and Olszowiec, Pawel and Byra, Michal and Stefanski, Grzegorz and Presta, Alberto},
  journal={arXiv preprint arXiv:2608.04879},
  year={2026}
}

@article{schonemann1966generalized,
  title={A generalized solution of the orthogonal procrustes problem},
  author={Sch{\"o}nemann, Peter H},
  journal={Psychometrika},
  volume={31},
  number={1},
  pages={1--10},
  year={1966},
  publisher={Cambridge University Press}
}

@inproceedings{zhang2022minivit,
  title={Minivit: Compressing vision transformers with weight multiplexing},
  author={Zhang, Jinnian and Peng, Houwen and Wu, Kan and Liu, Mengchen and Xiao, Bin and Fu, Jianlong and Yuan, Lu},
  booktitle={2022 IEEE/CVF Conference on Computer Vision and Pattern Recognition (CVPR)},
  pages={12135--12144},
  year={2022},
  organization={IEEE}
}

@inproceedings{shen2022sliced,
  title={Sliced recursive transformer},
  author={Shen, Zhiqiang and Liu, Zechun and Xing, Eric},
  booktitle={European Conference on Computer Vision},
  pages={727--744},
  year={2022},
  organization={Springer}
}

@article{johnston2023abstract,
  title={Abstract representations emerge naturally in neural networks trained to perform multiple tasks},
  author={Johnston, W Jeffrey and Fusi, Stefano},
  journal={Nature Communications},
  volume={14},
  number={1},
  pages={1040},
  year={2023},
  publisher={Nature Publishing Group UK London}
}

@article{loshchilov2017decoupled,
  title={Decoupled weight decay regularization},
  author={Loshchilov, Ilya and Hutter, Frank},
  journal=iclr,
  year={2019}
}

@article{vaswani2017attention,
title={Attention is all you need},
author={Vaswani, Ashish and Shazeer, Noam and Parmar, Niki and Uszkoreit, Jakob and Jones, Llion and Gomez, Aidan N and Kaiser, Lukasz and Polosukhin, Illia},
journal=neurips,
year={2017}
}

@string{cvpr={Proceedings of the IEEE Conference on Computer Vision and Pattern Recognition (CVPR)}}

@string{iclr={Proceedings of the International Conference on Learning Representations (ICLR)}}

@string{neurips={Advances in Neural Information Processing Systems (NeurIPS)}}

@string{springer={Springer International Publishing}}

@string{arxiv={{A}r{X}iv e-print}}

@misc{byra2026bvitinvestigatingsingleblockrecurrence,
      title={bViT: Investigating Single-Block Recurrence in Vision Transformers for Image Recognition}, 
      author={Michal Byra and Pawel Olszowiec and Grzegorz Stefanski and Grzegorz Gruszczynski and Alberto Presta},
      year={2026},
      eprint={2605.10661},
      archivePrefix={arXiv},
      primaryClass={cs.CV},
      url={https://arxiv.org/abs/2605.10661}, 
}
}
\end{document}